# A Study in a Hybrid Centralised-Swarm Agent Community

*Bradley van Aardt, Tshilidzi Marwala*

School of Electrical and Information Engineering
University of Witwatersrand
Johannesburg, South Africa.

b.vanaardt@ee.wits.ac.za  t.marwala@ee.wits.ac.za

**Abstract**

This paper describes a systems architecture for a hybrid Centralised/Swarm based multi-agent system. The issue of local goal assignment for agents is investigated through the use of a global agent which teaches the agents responses to given situations. We implement a test problem in the form of a Pursuit game, where the Multi-Agent system is a set of captor agents. The agents learn solutions to certain board positions from the global agent if they are unable to find a solution. The captor agents learn through the use of multi-layer perceptron neural networks. The global agent is able to solve board positions through the use of a Genetic Algorithm. The cooperation between agents and the results of the simulation are discussed here.

## 1. Introduction

Multi-agent systems are an engineering paradigm that has been gaining momentum over the past years [11]. A particular form of Multi-Agent systems, Swarm based systems have been successfully applied to a number of problems [2]

A large part of the difficulty in designing such systems is assigning Local strategies to individual agents in a community, with the result of some desired overall or Global behaviour [12]. In fact, the subject of assigning local strategies is of current research, and practical, interest. A methodology whereby agents are equipped with a host of strategies at design time, and then learn through interaction the appropriate local strategy to use in a given situation has been proposed by [16].

The ideal situation would be where agents are able to dynamically develop their own local behaviours, with the outcome of the given user desired Global behaviour, in response to the environment in which they operate. A possibility would be for agents to identify certain situations using pattern recognition techniques, and then act upon these situations using learned responses.

We present a study in this paper whereby agents learn response to situations, or local strategy, from an entity with a global view of the problem. In this way we theorise that the agents will learn local strategies that are likely to lead to the overall globally desired behaviour of the system. Firstly, we give a short background to the domain of Multi-Agent systems, and of the machine learning and evolutionary technology we use in the study. We then present the details of the study, and a brief analysis of our findings.

## 2. Background

### 2.1 Multi Agent Systems

The multi-agent paradigm, in which many agents operate in an environment, has become a useful tool in solving large scale problems through a "divide and conquer" strategy [7]. The Multi-agent system is a distributed, decentralised system. The paradigm of individual entities collaborating to solve a particular problem that is beyond each entities own capabilities is a natural concept, and one that is proving to be very powerful in practice [8]. However, while the concept is easily understandable, the implementation is not trivial. There are many complexities and subtleties in these such as [7]:

- Decomposing and allocating problems to the agents
- Describing the problem to the agents
- Enabling communication and interaction among the agents

Decentralised systems, in the context of Multi-Agent systems, promise the following advantages [7][9][3][4][10]

- No single failure point, therefore greater robustness. Multi-agent systems have the capacity to degrade gracefully.
- Possibility of faster response times and fewer delays as the logic/intelligence is situated nearer to the problem domain.
- Increased flexibility to take account of changes occurring in the problem domain.
- Modularity and Scalability. Multi-agent systems can be increased in size dynamically according to the demands of the problem.

The problems associated with Multi-agent systems are [4][2][7][11]:

- Difficulty in measuring and evaluating the stability and security of the system.
- Excessive communication between agents can slow down the system. This is often countered by heavily restricting the amount of communication between agents.
- Possibility of getting stuck in non-optimal solutions of the problem, often due to the lack of global knowledge of the problem from each agent's point of view.
- Most Multi-agent systems are built in an ad-hoc way since there is no absolute theory for these types of systems. There have been recent attempts however, to formalise the design of agent based systems, such as the Gaia Methodology [9]. However these are not yet in widespread use.

### 2.1.1 Swarm Based systems

Swarm intelligence is a particular paradigm for multi-agent systems which emphasizes distributedness and agent simplicity. It is based on the observations of social insects in nature, such as ants, termites and bees [2]. Such insect societies are extremely organized, even though there is no central control or planning. Each agent in the system is programmed only to achieve its own Local Goal. The agent's behaviours in Swarms are very simple: the intelligence of the system is 'emergent' from the overall behaviours of all the agents in the system. The communication between agents is usually performed indirectly [3], by agents making changes to the environment, which other agents act upon. This is analogous to insects laying pheromone trails to food source, etc. Emphasis is therefore placed on reactivity in these systems. Swarm systems have been successfully applied to many problems, notably routing in computer and telecoms networks [2] and recently to a manufacturing control system [3].

The disadvantage of swarm based systems is that no agents actually have a global view of the problem to be solved. All agents are entirely focused on achieving their own Local Goals, whether or not these goals are to the benefit or detriment to the overall community. This can exacerbate the problem of the system getting stuck in local optima, or worse, cause the system to fail.

The advantage of a centralised system, where there is effectively one agent or processing unit to control the whole system, is that such a system will make decisions which will benefit the Global Goal of the system.

### 2.2 Neural Networks

Artificial neural networks are inspired by the functioning of biological neurons in animal and human brains. Artificial neural networks differ from most other computing techniques in that they are able to learn arbitrary relations between sets of data presented to them. That is, rules are not explicitly programmed or set, but are learned from experience by the network [4]. The basic architecture of a neural network is also based on that of their biological counterpart: both networks consist of many simple elements, known as neurons, operating in parallel [5][6]. The most widely used neural network models are the Multi-layer perceptron (MLP) and Radial Basis Function (RBF) networks.

### 2.3 Genetic Algorithms

Genetic Algorithms (GAs) are inspired by the process of natural selection and evolution in nature. GAs are a class of non-greedy algorithms – they perform an exploration of the search space. GAs were first proposed by John Holland in 1975 [19] and can be thought of as four modules: *Encoding, Selection, Recombination and Evaluation*. Of the modules, only the Evaluation Function is problem specific. With a GA, possible solutions to a problem are encoded as a

*chromosome* string. A population of chromosomes make up a generation of possible solutions to the problem. From each generation, the fitness of each chromosome is evaluated by the problem-specific Evaluation module. Certain chromosomes are selected by the selection module, by some procedure, and the selected chromosomes are the *mated* by the Recombination module to form the next generation of chromosomes.

## 3. Hypothesis of study

Reference [1] states that the method of human learning when presented by a new task is to use rational reasoning to perform the decision making process behind solving the task. After using this deliberative approach, and we begin to "master" a task, we are able to perform it "naturally", without explicitly performing rational reasoning. At this point, people use their pattern recognition skills to perform the task.

It is hypothesised that the human model of discovery and learning can be applied to agent strategising. Evolutionary optimisation and genetic programming techniques can be used to discover strategies for the agent community. These methods are desirable due to the large number of possible solutions to the problem. This will be the deliberative stage, where a problem solution is formulated. Once a strategy has been discovered and used, a neural network can be trained with the results. This is the Pattern Recognition stage. Thus over time, the neural networks should learn a large number of strategies, while generalising these strategies to other scenarios. Thus, as agents perform more tasks, the problem solving should become "natural" to them, as the neural networks start taking over from the genetic programming and optimisation modules.

## 4. Method

Our paper is to link the qualities of the Swarm based (or highly decentralised system) with a Centralised system. To do this, we propose a system that consists of both a Centralised control, and a swarm of agents. Our method aims to use the Centralised control in the deliberative phase of problem solving, and then switch to a distributed system once the problem solutions, as discovered by the Centralised system, have been learned by the subconscious of the agents.

### 4.1 Test Problem

The proposed system is applied to the game of Pursuit. A Pursuit board is represented by a 2 dimensional grid pattern, as shown in Figure 1. The asterisks represent the Captor agents, while the circle represents the Fugitive agent (top row). The aim of the game is for the Captors to surround or corner the Fugitive. Each agent may only move one block per turn. Legal moves on a Pursuit board are Up, Down, Left, Right and Stay. No diagonal moves are allowed. At least one captor agent must move during the captors' move. Furthermore, the board does not wrap around. A game is ended when the fugitive is captured, implying that the fugitive has no place to move to.

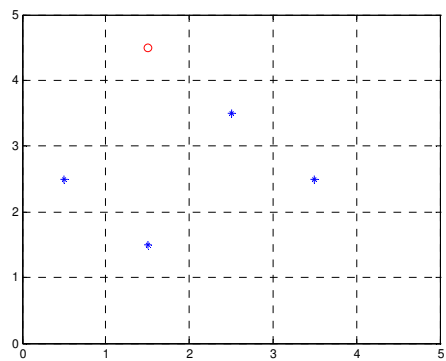

**Figure 1 Pursuit Board representation**

### 4.2 System Description

The architecture of the system is to have four separate agents representing the captors. Each of the agents has a neural network, which are trained independently of each other. Each agent is responsible for making a valid move.

The system also has a Genetic Algorithm based agent, which can control all of the Captor agents, called the Global Agent. It has a global view of the system. When the Global Agent is invoked, it acts as a centralised system, which proposes moves that best satisfy the Global goal of the system.

The system operation is shown in flowchart form in Figure 2. At each turn for the captors to move, the Captor agents are invoked. The proposed moves from each of the captors are then combined, and the result is tested for legality. If the proposed moves for all the captors are legal, the moves are implemented synchronously. If not, the Global Agent is invoked, and the moves proposed by the Global Agent are implemented. The new move data is then used to re-train the Captor Agents.

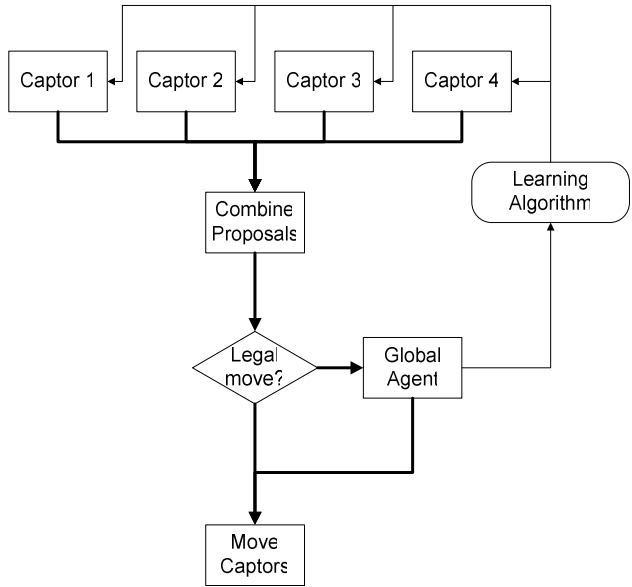

**Figure 2 System Flow Diagram**

The fugitive agent has no real intelligence; it merely chooses a (legal) move at random.

The system is thus a hybrid Centralised/Swarm system. The centralized system is effectively used to train the multi-agent system. The advantage of this arrangement is that the centralized system has a global view of the Pursuit board. The Centralised system, as represented by the Global Agent is programmed to favour a solution that minimises the sum of all the captor's distances from the fugitive. In other words, it looks for an arrangement of the captors that will maximally surround the fugitive. If a pure swarm system approach were to be followed, a likely implementation would be one where each agent attempts to minimize its own distance from the fugitive. This will cause agents to get in each others way, by blocking, or will result in an inferior strategy by the agents, since they are not prepared to make sacrifice to their positions, even if it would result in a globally better arrangement of the captors.

Initially, as the system is run, it operates mainly as a Centralised system, with the Global Agent proposing most of the moves. As time progresses, however, the Captor agents start performing more of the moves and the system edges towards a swarm system.

### 4.3 Captor Agents Description

The agents each contain a neural network. The network inputs are the current relative positions of all other agents (captors and fugitive) on the board to the agent. There are five outputs for each network, each representing a particular direction in which the agent should move.

#### 4.3.1 Board Representation and Training

The input representation to a neural network is extremely important to the success of practical network applications [13][14][15]. Smooth representation of the input data, as well as using an input representation that describes *features of* the data can help reduce the number of states the network has to learn in a lookup-table type fashion, and increase the ability of the network to generalise learned data to new situations [15].

For this study, we implement the board representation as a set of relative differences in position. Each Captor agent has the relative positions of the other captors inputted in order of Cartesian distance. The relative position of the fugitive is then given as the last two inputs. This is depicted in Figure 3.

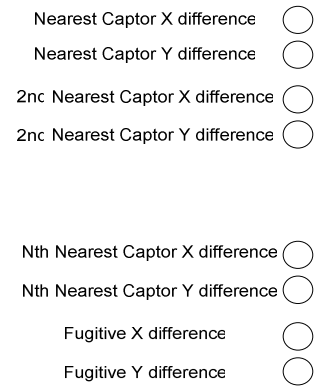

**Figure 3 Network Input representation**

This representation allows various formations of the agents to be captured, regardless of the exact location on the Pursuit board. Such formations are common when the captors are chasing the fugitive.

The Captor Agents are trained on moves discovered by the Global Agent, using its Genetic Algorithm. Each Captor is trained with data specific to its own experience on the board. Training the agents from a centralised system means that the direct communication between agents can be drastically reduced. This is due to the fact that each agent will take into account other agents moves when moving itself. Therefore, in terms of communication, the system acts as a swarm system, but with one major difference: each agent is "aware" of the global goal through the training it has received from the Global Agent, and so inherently acts cooperatively with other agents.

### 4.4 Global Agent Description

The evaluation function for a chromosome of moves is calculated by summing the Euclidean distance of each of the Captor agents from the Fugitive agent. This is given by:

$$S = \sum_{N=1}^{M} \sqrt{(X_N - X_A)^2 + (Y_N - Y_A)^2}$$

Where: S is the sum of the distances to the fugitive; M is the number of Captor agents in the system; $X_A$, $Y_A$ are the Fugitive agent's coordinates and $X_N$, $Y_N$ are the current Captor's coordinates.

In order to ensure that illegal moves are discouraged, a penalty scheme is used where the sum is multiplied by a large number in order to create a peak in the evaluation function. Similarly, to ensure that a winning move (when possible) is chosen clearly above any other possible legal combination, such as move is heavily rewarded.

### 5. Implementation

The system is implemented in MATLAB. The Genetic Algorithm toolbox used is the open source GAOT [18]. NETLAB [17] is used to implement the neural networks for the Captor Agents. A two-layer MLP architecture is used for the networks. The networks use a Logistical function for the output units. A variable number of hidden units are used, depending on the amount of data the network has seen.

Since NETLAB does not directly support on-line training, an alternative scheme is used. Every time the Global Agent is invoked, the board position is recorded, as well as the moves proposed by the Global Agent. This data is added to the set of training data. After each game is complete, the agents are re-trained using the updated training data.

### 5.1 Discussion

Despite the fact that Pursuit is a very simple game, the successful suggestion of moves by the community of agents is quite significant. Firstly, to make a legal move, all the agents have to be aware of the rules of the game, which has not been explicitly programmed into them. Secondly, successful moves invariably require some degree of cooperation with neighbouring agents. Usually, this is achieved by explicitly using communication between agents. In this system, there is no such communication. Successful moves mean that the agents are able to anticipate each other's actions.

After a large number of games, each agent will have observed a large variety of board positions, and for this system, the neural networks will tend to be the same. This is only because the agents in this system all have the same "abilities" or allowed moves. In a scenario where there are agents with different abilities, as for the separate pieces in chess, the networks will each train on totally different output moves, and so will tend to be different even after observing an infinite number of moves. However, there is no reason why the agents should not still be able to anticipate each others moves, since they will be trained with this data.

### 6. Recommendations and Further work

On many of the occasions, the agents were not able to make a legal move due to the fact that some agents suggested more than one direction to move in at a time. While this intuitively suggests that the agents are not yet well trained in the particular situation, this is often not the case. The genetic algorithm does not always propose the same result for a particular board position. This means that for a given board position, an agent may be trained with many possible moves. What is needed is a method to resolve the proposed multiple directions by an agent in a particular situation with the other agents.

Further work involves a methodology allowing the agents themselves to formulate a global strategy, without relying on the external Global agent. This could be a simple principle such as distributing the evaluations of the Genetic Algorithm populations among the agents, or more elaborate methods.

## 7. Conclusion

The study of a hybrid Centralised-Swarm agent community has been discussed in this paper. The implementation of a proposed architecture, in the form of a Genetic Algorithm as the Centralised/Global agent, and as neural networks as swarm agents, is applied to the game of pursuit.

We found that although the swarm, or Captor, agents are able to learn a large number of moves, the fact that more than one move may be applicable in a situation results in agents proposing all these possibilities in a situation. This needs to be resolved within the swarm system to result in a valid move.

The possibility of reduced communication between agents, while still cooperating with each other is demonstrated here.